\documentclass[conference]{IEEEtran}
\usepackage{ctable}
\usepackage[linesnumbered,vlined,ruled,boxed,commentsnumbered]{algorithm2e}
\usepackage{bm}
\usepackage{amsmath}
\usepackage{cleveref}
\usepackage{subfigure}
\usepackage{cite}
\usepackage{comment}
\newcommand{\fref}[1]{Fig. \ref{#1}}
\newcommand{\tref}[1]{Table \ref{#1}}

\newcommand{\sref}[1]{Section \ref{#1}}
\newcommand{\eref}[1]{Eq. \ref{#1}}
\newcommand{\aref}[1]{Alg. \ref{#1}}

\usepackage{xcolor} %
\begin{document}

\title{Self-Organizing Maps with Variable Input Length for Motif Discovery and Word Segmentation}

\author{\IEEEauthorblockN{Raphael C. Brito, \textit{Member, IEEE}, and Hansenclever F. Bassani, \textit{Member, IEEE}}
\IEEEauthorblockA{Center of Informatics - CIn,
Universidade Federal de Pernambuco, Recife, PE, Brazil, 50.740-560\\
Email: \{rcb7,hfb\}@cin.ufpe.br}
}

\maketitle

\begin{abstract}
Time Series Motif Discovery (TSMD) is defined as searching for patterns that are previously unknown and appear with a given frequency in time series. Another problem strongly related with TSMD is Word Segmentation. This problem has received much attention from the community that studies early language acquisition in babies and toddlers. The development of biologically plausible models for word segmentation could greatly advance this field. Therefore, in this article, we propose the Variable Input Length Map (VILMAP) for Motif Discovery and Word Segmentation. The model is based on the Self-Organizing Maps and can identify Motifs with different lengths in time series. In our experiments, we show that VILMAP presents good results in finding Motifs in a standard Motif discovery dataset and can avoid catastrophic forgetting when trained with datasets with increasing values of input size. We also show that VILMAP achieves results similar or superior to other methods in the literature developed for the task of word segmentation.

\end{abstract}

\begin{IEEEkeywords}
self-organizing maps, variable input length, subspace clustering, motif discovery.
\end{IEEEkeywords}

% \fi
\IEEEpeerreviewmaketitle

\section{Introduction}

Motifs are described in the literature as recurrent patterns, frequent tendencies, successions, forms, episodes, or frequent subsequences that occur in time series\cite{torkamani2017survey}. Motif Discovery methods search for previously unknown frequent patterns in a time series \cite{lin2002finding}. The task can be seen as a time series clustering problem, assuming that each cluster must group together patterns in the time series that represent the same Motif. The problem of Word Segmentation in transcription of fluent speech can be seen as a Motif discovery problem in which the words are the Motifs of a time series composed of phonemes. This problem has received much attention from the community that studies early language acquisition in babies and toddlers and is one of the focus of the present article. The development of biologically plausible models for word segmentation could greatly advance this field.

The Self-Organizing Map (SOM) \cite{Kohonen1998} is a biologically inspired neural network, frequently applied for visualizing high dimensional data by compressing the information while preserving the topological relations and capturing the most important characteristics of the input data. However, it finds applications in many other types of problems, such as surface reconstruction \cite{Rego2007} and data clustering \cite{Bassani2015}. It is also worth mentioning that SOM has been applied to a variety of problems involving sensory processing, including visual \cite{Miikkulainen2005} and auditory information \cite{Kangas1991}.
%http://ieeexplore.ieee.org/document/150288/
%http://ieeexplore.ieee.org/document/150288/
The original SOM \cite{kohonen1982self} defines a map composed of a set of nodes, or prototypes, that compete and cooperate to represent a certain region of the input space. The nodes are usually organized in a fixed bi-dimensional grid and the model employs an Euclidean distance for comparing the input patterns with each node in map, what may not be adequate for clustering high-dimensional datasets, due to the curse of dimensionality \cite{keogh2011curse} or in datasets in which not all dimensions are equally relevant for the different clusters, such as in subspace clustering \cite{reviewClustering2004, clusteringSurvey2009} and Motif discovery. However, new models based on SOM were developed for improving its performance in such scenario \cite{DSSOM,Bassani2015}.

The Local Adaptive Receptive Field Dimension Selective Self-organizing Map (LARFDSSOM), proposed in \cite{Bassani2015}, is an example that presented good results for the task of subspace clustering. The model achieves this by applying different weights for each input dimension for each cluster. These characteristics enable a new range of clustering related applications beyond subspace clustering, in which Motif Discovery is included. However, in Motif discovery, there is a demand for methods that can work with different sizes of samples \cite{nunthanid2011discovery} in order to allow the discovery of Motifs with different lengths. LARFDSSOM was not developed for this case and to the best of our knowledge, there is no SOM-based method for clustering data with supporting inputs with different length.

In addition to that, the Variable Input Length MAP (VILMAP) was developed to extend LARFDSSOM to make it possible clustering patterns with different sizes. VLIMAP takes advantage of the self-adjustable input weights of LARFDSSOM for allowing the input samples with a variable regarding the number of input dimensions. Therefore, when VILMAP is trained with patterns of different sizes, it generates a map with prototype nodes that have different sizes.

In our experiments, we show that VILMAP is able to find Motifs in a standard Motif discovery dataset (the GunPoint dataset \cite{UCRArchive}). Additionally, we verified that VILMAP avoids catastrophic forgetting when trained with datasets with increasing input sizes. Finally, we also show that VILMAP achieves results similar or superior to other methods in the literature developed for the task of word segmentation.

The rest of this paper is organized as follows: Section \ref{sec:related_works} presents the related work; Section \ref{sec:variable_input_length_map} describes the proposed model; Section \ref{sec:experiments} presents the experimental setup and the obtained results; and finally, in \sref{sec:conclusion}, we present our final considerations.

\section{Related Work}\label{sec:related_works}
In the \sref{Motif_discovery} a description of Motif Discovery is presented with a method of the area; already in \sref{som_Motif}, will be explained how the Self-Organizing Map works with Motif Discovery problem; and finally, in the \sref{word_segmentation} four methods that are applied to the word segmentation area will be displayed.

\subsection{Variable Motif Discovery}\label{Motif_discovery}

A time series (TS), $S$ can be defined as a list $\quad S\quad = \quad ({ s }_{ 1 },\quad { s }_{ 2 },\quad...,\quad s_{ n })$ of real-valued variables, where $n$ is the total size of the series, that is, the number of points in the series. Motifs are called the frequent patterns in a time series, which are previously unknown, while the search for these patterns is called Motif Discovery \cite{lin2002finding}.

Defining $R$ as a threshold that establishes a minimum allowed similarity or the maximum distance allowed between two occurrences of a Motif, there are two major definitions for the problem of Time Series Motif Discovery (TSMD), according to Mueen (2014) \cite{mueen2014}:

\noindent
\textit{Similarity-based Definition}: Given a time series and its length, the time series Motifs are the segments repeated in the order of their similarities between repeated occurrences within a range $R$.

\noindent
\textit{Support-based Definition}: Given a time series and its length, time series Motifs are the segments that have the most number of repetitions within a range $R$.

The Variable-Length Motif Discovery (VLMD)\cite{nunthanid2011discovery} is a method that has been proposed to automatically find a suitable set of variable length Motifs. VLMD iteratively separates Motifs of different sizes into groups based on similarity. Within a group, a representative Motif with a normalized minimum distance between pairs of subsequences is selected. Finally, the VLMD returns a set of useful representative Motifs, which is extremely small compared to all possibilities of sliding window lengths. The operation of VLMD consists of two steps. Firstly, it finds a set of groups of Motifs looking for all possible lengths of a sliding window to obtain Motifs of different lengths. If the current Motif and the previous Motif overlap, the current Motif is added to the same set of the previous Motif; Otherwise, a new Motif group is created. Then, for each group of Motifs, a representative Motif is selected with a minimum normalized distance for the others in the group.

Therefore, the VLMD can return a small set of motifs from a given time series sequence; this method does not present an application in Word Segmentation area, and we not compare with this method because \cite{nunthanid2011discovery} did not provide enough detailed explanation to reproduce the experimental results.

\subsection{Self-Organizing Maps for Motif Discovery}\label{som_Motif}

\begin{figure}[ht]
	\centering
	\includegraphics[width=0.90\linewidth]{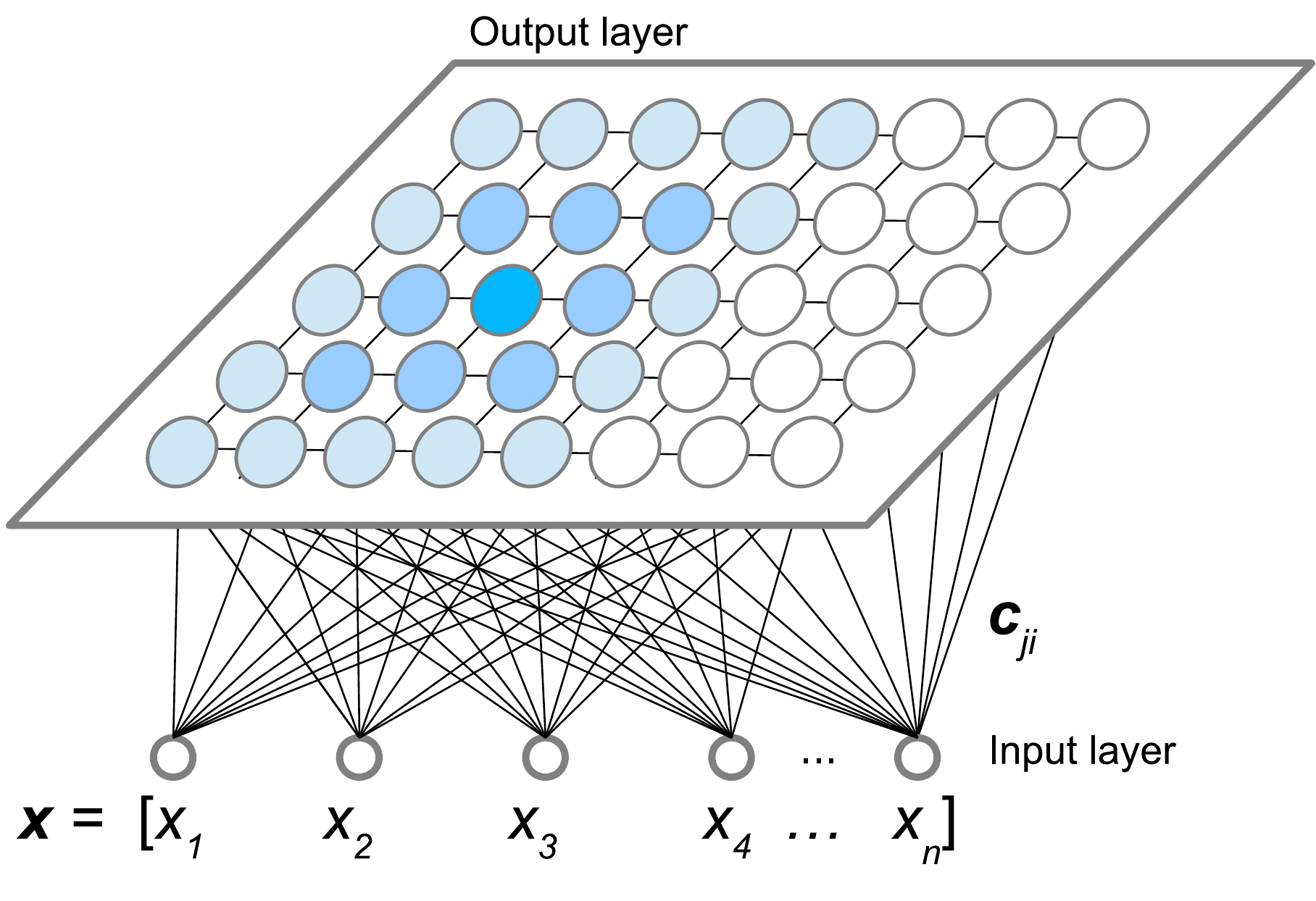}
	\caption{The basic structure of a SOM. Where $\textbf{x}$, is the input pattern, $x_i$ is the values of the $i$-th input layer node. $\textbf{c}_{ji}$, represents weight between with the $j$-th node in the output layer (organization layer) with the $i$-th node in the input layer. In this configuration, each node in the output layer is directly connected with four neighbors on the rectangular grid.}
	\label{fig:SOMstructure}
\end{figure}

The Self-Organizing Map (SOM) (\fref{fig:SOMstructure}), proposed by \cite{kohonen1990self}, is a neural network that maps a high-dimensional data into a smaller, usually bi-dimensional grid of $N$ nodes (or neurons), compressing information while preserving the topological relationships of the original data. The nodes in the grid, which position in the input space is $\textbf{c}_{j=1..N}$, participate in a winner-takes-all competition to represent each input stimulus, $\textbf{x}$. The nearest node to an input stimulus, $\textbf{c}_{s}$ (the winner node), is slightly moved to approximate the $\textbf{x}$. The neighbors of the winner node in the grid are adjusted as well (cooperation), in a smaller scale, to establish a topological relationship in the grid that reflects what is observed in the input space. After training, similar stimuli are clustered in the same node on the grid or in topologically near nodes.

LARFDSSOM is a model based on SOM that has a time-variant structure, in which nodes are inserted or removed from the map during the self-organization process, whenever it is required for to better representing the patterns in input space. Also, in LARFDSSOM, nodes can apply different relevances for each input dimension and perform the adjustment of the receptive fields as a function of the local variance observed in the input data.

	The operation of the map takes place in three phases: the organization phase, the convergence phase, and the clustering phase \cite{Bassani2015}.  In the organization phase, nodes are dynamically inserted, removed and adjusted in the map in order to cover the regions of the input space in which the input patterns are found as well as possible. When an input pattern is presented for the network, a level of activation is computed for each node in the map, and the node with the highest activation is considered the winner of the competition. This activation level is an inverse function of the distance between the center of the node and the input pattern. An activation threshold, $a_t$, is used for determining when if the winner node is close enough to be adjusted or if it is necessary to insert a new node in the map. The nodes that do not cluster a significant percentage of the input patterns are periodically removed.

    In the convergence phase, the nodes are also updated and removed when required, similar as in the organization phase. However, there is no insertion of new nodes. This phase finishes when the number of nodes in the map stops decreasing.

    After convergence phase, the information stored on the map can be used for clusters new input patterns (clustering phase). All nodes with an activation equal to or greater than the threshold $a_t$ for a specific input pattern are considered as clustering it.

    Finally, as well as in SOM, the input layer of LARFDSSOM has a fixed length, thus, it does not allow the increase or decrease of the sample sizes during the self-organization process.

\subsection{Word Segmentation}\label{word_segmentation}

Unlike the spacing provided in the written text, the spoken words are rarely delimited by pauses, so children must learn somehow to identify the boundaries between words as they hear the sentences. Because the structure of words is significantly variable in all languages, it is difficult to know how a child between 9 and 15 months of age achieve this ability. Therefore, segmentation is a key step in language acquisition, specifically in lexical development \cite{saffran1996word}. It is worth pointing out that finding the exact boundaries of all words in a sentence is not strictly necessary for children to understand what is said. They actually need to recognize the words present in the sentence (simple or compound) in the correct order, what can be done despite the occurrence of certain boundary detection errors, as it is observed in young children \cite{Correa2007}.

In the present work, we reproduce the experiments described in \cite{Larsen_Cristia_Dupoux_2017} and compared the results we obtained with the following four algorithms of Word Segmentation:

\subsubsection{DiBS}
Diphone-Based Segmentation (DiBS) \cite{daland2011learning} is based on phonetic properties and keeps in memory the frequency of two phonemes that occur together and decide to place a boundary between them by computing Bayes' theorem. To do this, the model implies certain assumptions: the learning algorithm knows the phonetic categories, it is able to detect expression boundaries, it assumes phonological independence through word boundaries, it traces the free distribution of context of the diphones and knows the relative frequency of word forms already learned. Therefore, it uses the local statistical clues in order to determine the word boundaries.

\subsubsection{TPs}
Another local statistical model is based on tracking Transitional Probabilities (TPs) \cite{saffran1996statistical} over syllables which posit a boundary between two syllables if its co-occurrence probability is locally lowest (relative threshold) or is lower than an absolute threshold, usually computed by taking the averaged value of syllables pairs. TPs demands a larger memory as the number of all possible syllables encountered are much greater than the number of all possible phonemes if compared with DiBS.

\subsubsection{PUDDLE}
The algorithm Phonotactics from Utterances Determine Distributional Lexical Elements (PUDDLE) \cite{monaghan2010words}, incrementally creates a lexicon using information about expression limits and deducing phonetic constraints. More precisely, each time a sequence of phonemes is found if correspondence with a word in the proto-lexicon is found and if certain phonological constraints are respected, the chunk of phonemes is added to the proto-lexicon. The initial phoneme pairs and endings are added, respectively, to lists of pairs of initial and final phonemes previously encountered.

\subsubsection{AGu}
Unigram Adapter Grammar (AGu) \cite{johnson2007adaptor} models an ideal learner, that is, a learner who has an infinite memory and a batch process, observing the whole corpus before segmenting. The structure consists of two modules: a lexical generator and an adapter. The first generates a lexicon of items that are likely to be found in the corpus and the second assigns item frequencies. Importantly, unigram AG assumes that lexical items are generated independently of each other and that the stochastic process is chosen so that the frequencies of the items follow a power law distribution as found in natural language.

In the next section, the proposed method (VILMAP) will be presented aiming at addressing the problems of word segmentation.

\section{Variable Input Length Map}\label{sec:variable_input_length_map}
VILMAP\footnote{Available at: https://github.com/RaphaBrito/VILMAP} is a model based LARFDSSOM \cite{Bassani2015} that is capable of receiving stimuli from an unknown number of dimensions. The proposed model inherits some important features from LARFDSSOM: 1) The first one is the ability of the model to adapt its structure as new patterns are presented over time; 2) The second feature inherited by VILMAP is the capacity that the nodes have to apply different relevances for each input dimension; 3) the receptive field of the nodes can be adapted during the self-organizing phase of the model.

The VILMAP is composed of two phases: Self-organization phase \aref{alg:somPhase} and Clustering phase \aref{alg:clusteringPhase}. The convergence phase that exists on the map was not inherited for VILMAP because we aim to work in an online way. As it occurs in the standard SOM, the organization phase of VILMAP is composed of three steps: 1) competition, 2) adaptation, and 3) cooperation. Similarly, as in LARFDSSOM, when a new input pattern is presented, a competition occurs among the nodes, and the winner is the most active node according to a radial base function (see line 5 of the \aref{alg:somPhase}). However, a shift operation is employed to compare the input pattern signal with the information stored on the prototype, the shift operation verifies the input pattern signal in all possible position shifted. After this process, a winner node will be obtained, to be the chosen, it needs to have the highest value in one of his shifts. Then, it is verified if its activation is above a threshold parameter, $a_t$, when it happens,  a new node is created with its center at the same position as the input pattern. On the other hand, if the node activation is greater than $a_t$, the adaptation and cooperation steps take place: the winner node is updated to approximate of the input pattern (adaptation) as well as their neighbors (cooperation). This procedure is detailed in \sref{updating_winner_node_and_neighbors}.

After one execution of the organizing phase, the clustering phase can be performed. In VILMAP the clustering procedure associates the input pattern to only one cluster, and it is represented by the node on the map that returned the highest activation. The clustering procedure is detailed in \sref{projected_clustering_VILMAP}.
%1)Tente manter uma padronizacao no seu texto, no uso de nomes como pattern, uso de maiusculas. Tente deixar o texto mais claro e objetivo.
%2)Enumere as fases quando for necessario, isso da uma ideia de um passo a passo do processo de aprendizagem do algoritmo e deixa mais claro para o leitor como o algoritmo funciona.
%3)Evite usar o google translator sem verificar se a palavra sugerida eh adequada para o texto. O google translator e um ferramenta muito boa, mas ela deve ser apenas utilizada como um guia.
%4)Verifique se as suas referencias estao corretas.

\begin{algorithm}[!ht]
 Initialization of parameters ${ a }_{ t },$\quad ${ e }_{ b },$\quad ${ e }_{ n },$\quad $\beta,$\quad ${\epsilon}_{ds},$\quad ${ N }_{ max },$\quad $D_{ min },$\quad ${ D }_{ max }$\quad $and$ \quad $minwd$ \;
 
 Initialize the map with one node with ${ c }_{ j }$ initialized at the first input pattern, ${ \delta  }_{ j } \leftarrow 0$ and ${ \omega  }_{ s } \leftarrow 1$ \;
 
\While{have input pattern}{
    Input pattern \textbf{x} is presented to the Map\;
    The activation between \textbf{x} and all nodes is calculated according to \eref{eq:activation}\;
    Find the winner s with the highest activation (${ a }_{ s }$) conforming to \eref{eq:competition}\; 
    \If{$size(s) < size(x)$}{
    	updateNodeDimension(s, x) described in the \sref{sec:variable_input_length_map}\;
    }
     \uIf{${ a }_{ s } < { a }_{ t }$ and $N < { N }_{ max }$ }{
       Create new node $j$ and set: \boldmath$ { c }_{ j } \leftarrow x $ and \boldmath$ {\delta}_{j} 			\leftarrow 0$\;
       Connect $j$ to the other nodes\; 
       $N \leftarrow N +1$\;
     }\uElseIf{${ a }_{ s } \geq { a }_{ t }$}{
		Update the distance vectors \boldmath${ \delta }_{ s }$ of the winner node and of its 				neighbors\;
        Update the relevance vectors \boldmath${ \omega }_{ s }$ of the winner node and of its 				neighbors\;
		Update the weight vectors $\bf{ c }_{ s }$ of the winner and of its neighbors\;
     }
  
 }
\caption{Self-Organization Phase}
\label{alg:somPhase}
\end{algorithm}

\begin{algorithm}[!ht]
\While{have input pattern}{
    Input pattern \textbf{x} is presented to the Map\;
    The activation between \textbf{x} and all nodes is calculated  according to \eref{eq:activation}\; 
    Find the winner $s$ with the highest activation (${ a }_{ s }$) that is calculated using \eref{eq:competition} \; 
    \If{${ a }_{ s } \geq { a }_{ t }$}{
    	Assign \textbf{x} to the cluster with the index of the
winner node $s$\;
    }
     
 }
\caption{Clustering Phase}
\label{alg:clusteringPhase}
\end{algorithm}

\subsection{Competition and Insertion of Nodes}\label{competition_and_insertion_of_nodes}

Each node $j$ in the VILMAP represents a cluster associated with three $m$-dimensional vectors, where $m$ is the current number of input dimensions. 1) the first dimension is the center of the vector ${ c }_{ j } = \{ { c }_{ ji },  i=1, ..., m\}$ that represents the cluster prototype $j$ in the input space; 2) the second is the relevance vector ${ \omega }_{ j } = \{ { \omega }_{ ji },  i=1, ..., m\}$, which stores the weights (varying between $[0, 1]$) that the node $j$ applies to the $i$-th input dimension; 3) and the third is the distance vector ${ \delta }_{ j } = \{ { \delta }_{ ji },  i=1, ..., m\}$, which stores a moving average of the distances obtained from the input patterns \textbf{x} and the center of the vector $|$x$-{ c }_{ j }(n)|$, which is only used to compute the relevance vector as in \cite{Bassani2015}.

A radial basis function of a weighted distance \eref{eq:activation} is used to calculate the activation of a node in the VILMAP. The nodes receptive fields are adjusted as a function of the weighted distance among the prototype, the input pattern, and the summation of the relevance vector $\sum _{ i=1 }^{ m }{ { \omega  }_{ ji } }$.  Lower distances and higher relevances result in a higher activation (\eref{eq:activation}). The equation of activation is
\begin{equation}
ac({ D }_{ \omega  }(x, { c }_{ j }), { \omega  }_{ j })\quad =\quad \cfrac { \sum _{ i=1 }^{ m }{ { \omega  }_{ ji } }  }{ \sum _{ i=1 }^{ m }{ { \omega  }_{ ji } } + { D }_{ \omega  }(x, { c }_{ j }) + \epsilon  } 
\label{eq:activation}
\end{equation}

where $\epsilon$ is a very small value to avoid division by zero and ${ D }_{ \omega  }(x, { c }_{ j })$, is the weighted distance shown in \eref{eq:distance} as proposed in \cite{DSSOM}.

\begin{equation}
{ D }_{ \omega  }(x,\quad { c }_{ j })\quad =\quad \sqrt { \sum _{ i=1 }^{ m }{ { \omega  }_{ ji }{ ({ x }_{ i } - { c }_{ ji }) }^{ 2 } }  } 
\label{eq:distance}
\end{equation}

When a node is created, its center ${c}_{j}$ is initialized at the position of the last input pattern presented to the map. The relevance vector ${\omega}_{j}$ is initialized as an array of ones and the distance vector is initialized as an array of zeros. These vectors are updated in the adaptation and cooperation steps, presented in the \sref{updating_winner_node_and_neighbors}.

As in LARFDSSOM, the winner of a competition $s(x)$, is the node that presents the highest activation value related to the input pattern, as defined in the \eref{eq:competition}.
\begin{equation}
	S(x)\quad =\quad \underset { j }{ argmax } [ac({ D }_{ \omega  }(x,\quad { c }_{ j }),\quad { \omega  }_{ j })]
\label{eq:competition}
\end{equation}

VILMAP has an activation threshold ${a}_{t}$. If the winner node activation is below ${a}_{t}$, a new node is created at the same position of the input pattern and the winner node is not modified. Otherwise, when the winner node activation is above ${a}_{t}$, the winner node and its neighbors are updated, as described in the next section and shown in \aref{alg:somPhase}. Therefore, the activation threshold ${a}_{t}$ affects the final number of nodes in the map.

Since VILMAP can receive inputs with different sizes, it is necessary to develop a new procedure for computing the distance \eref{eq:distance} the input patterns with the nodes in the map.

\begin{enumerate}

\item \textbf{regular comparison}: if the node size in length is equal to the size of the input pattern, then the calculation of the distance is straightforwardly calculated as in LARFDSSOM (\eref{eq:distance}), with node and input pattern completely aligned.

\item \textbf{sliding window comparison}: If the winner node is greater in length than the input pattern, the information stored in the node will be compared with each part of the input pattern, as in a sliding window approach, shifting one position at a time. Then the activation is computed for each position according to the \eqref{eq:competition} and the highest activation value is returned as a result at the end of the process. This is illustrated in \fref{fig:activation_desloc} in which two displacements are possible when the length of the pattern is 4 and of the node is 5, thus, two activations are computed. In this example, the node achieves a greater activation with the displacement B.

\begin{figure}[!ht]
    \centering
    \includegraphics[scale = 0.3]{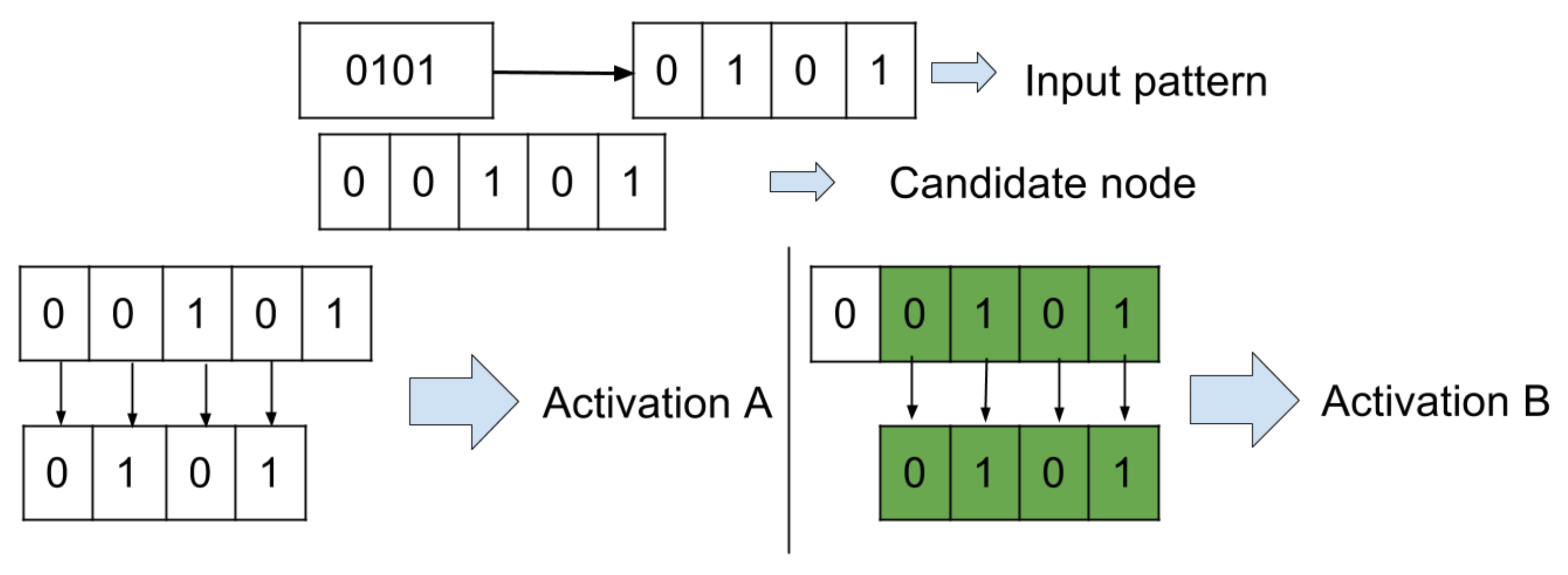}
    \caption{Example of how the activation is calculated in a binary dataset for an input pattern smaller than the node.}
    \label{fig:activation_desloc}
\end{figure}

\item \textbf{truncated comparison}: If the node is smaller in length than the input pattern, the activation is calculated only on the initial part of the input pattern and the final part of the stimulus is disregarded as illustrated in \fref{fig:add_dimension}.  In this case, if a node that has undergone through this procedure becomes a winner and has an activation above $a_t$, it will have its dimension updated: all vectors of the winner node grow to the same size as of the input pattern. Then, the new elements of the vector \boldmath${ c }_{ j }$ are initialized with the same values of the input pattern \textbf{x}; the new values of vector \boldmath${ \delta }_{ j }$ are initialized with 0.0; and the new elements of the vector \boldmath${ \omega }_{ j }$ are initialized with an intermediary value 0.5. \fref{fig:add_dimension} also illustrates such dimension update procedure.

\begin{figure}[!ht]
    \centering
    \includegraphics[scale = 0.29]{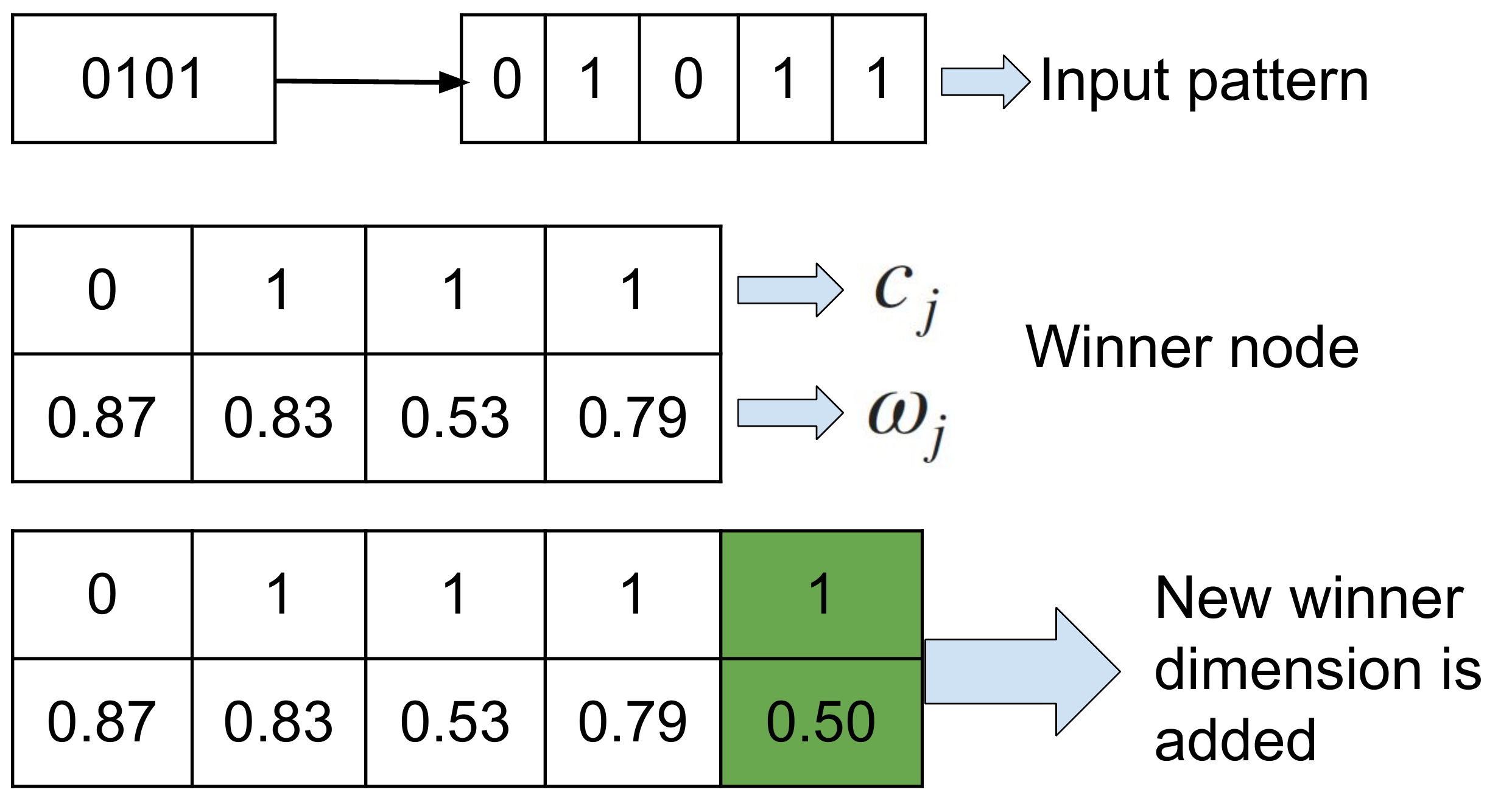}
    \caption{The example illustrating the node vectors dimension update.}
    \label{fig:add_dimension}
\end{figure}

\end{enumerate}

\subsection{Updating the Winner Node and Its Neighbors} \label{updating_winner_node_and_neighbors}

VILMAP can receive stimuli of different sizes at any moment. Hence, the $m$-dimensional vectors of the winner node might have a different size of the input pattern. Here again, three situations are possible:

\begin{enumerate}

\item \textbf{regular update}: if the node size in length is equal to the size of the input pattern, the node is updated straightforwardly as in LARFDSSOM, with node and input pattern completely aligned.

\item \textbf{update without the end}: If the winner node is higher in length than the input pattern, the update is done by first aligning the pattern with the first position of the winner node. Then, the vectors are updated as in LARFDSSOM, disregarding the parts that lie out of the vectors.
%\item \textbf{displaced update}: If the winner node is higher in length than the input pattern, the update is done by first aligning the pattern with the winner node precisely at the position that produced the highest activation. Then, the vectors are updated as in LARFDSSOM, disregarding the parts that lie out of the vectors.

\item \textbf{truncated update}: If the node is smaller in length than the input pattern, then, first, the dimensions of vectors associated with the node are update as follows: all vectors of the node grow to the same size as of the input pattern. Then, the new elements of the vector \boldmath${ c }_{ j }$ are initialized with the same values of the input pattern; the new positions of vector \boldmath${ \delta }_{ j }$ are initialized with 0.0; and the new elements of the vector \boldmath${ \omega }_{ j }$ are initialized with an intermediary value (0.5 assuming data normalized in 0-1 interval). \fref{fig:add_dimension} also illustrates such dimension update procedure. Finally, the aligned vectors are updated as in LARFDSSOM.

\end{enumerate}

\subsection{Clustering with VILMAP}\label{projected_clustering_VILMAP}
After the organization phase, the information stored in each node of the VILMAP can be used to cluster the test input patterns. Algorithm 2 presents the clustering procedure. Each node in the map defines one cluster and all test patterns for which a certain node is the winner are clustered together. In the clustering phase, the activation of the nodes is calculated in the same way as described in \sref{updating_winner_node_and_neighbors}.

\subsection{Setting Parameters for the Model} \label{Setting_Parameters_Model}
Parameter adjustment is performed as in LARFDSSOM. \tref{table:parameters} shows the ranges used for each parameter. The most important parameter for VILMAP is $ a_t $ because it directly influences the number of nodes that will be created. The smaller the $ a_t $, the fewer nodes will be inserted in the map, since with the high threshold the nodes will recognize fewer input patterns, causing more nodes created on the map.

\begin{table}[!ht]
\centering
\caption{Parameter Ranges for VILMAP}
\label{table:parameters}
\begin{tabular}{l|ccc}
\specialrule{.1em}{.05em}{.05em} 
Parameters & max & min  \\ \hline
Activation threshold ($a_t$)    & 0.70     & 0.999 \\
Relevance rate {$\beta$}    	& 0.001     & 0.5  \\
Winner learning rate ($e_b$)    & 0.0001    & 0.01 \\
Neighbors learning rate ($e_n$) &   0.002   & 1.0$\times e_b$  \\
Connection threshold ($minwd$)  & 0.001     & 0.5  \\
Relevance smoothness (${\epsilon}_{ds}$) &	0.01	&	0.1	\\
\specialrule{.1em}{.05em}{.05em} 
\end{tabular}
\end{table}
\begin{figure*}[!ht]
    \centering
    \hfill
    \subfigure[Procedure A]
    {\includegraphics[width=7.9cm]{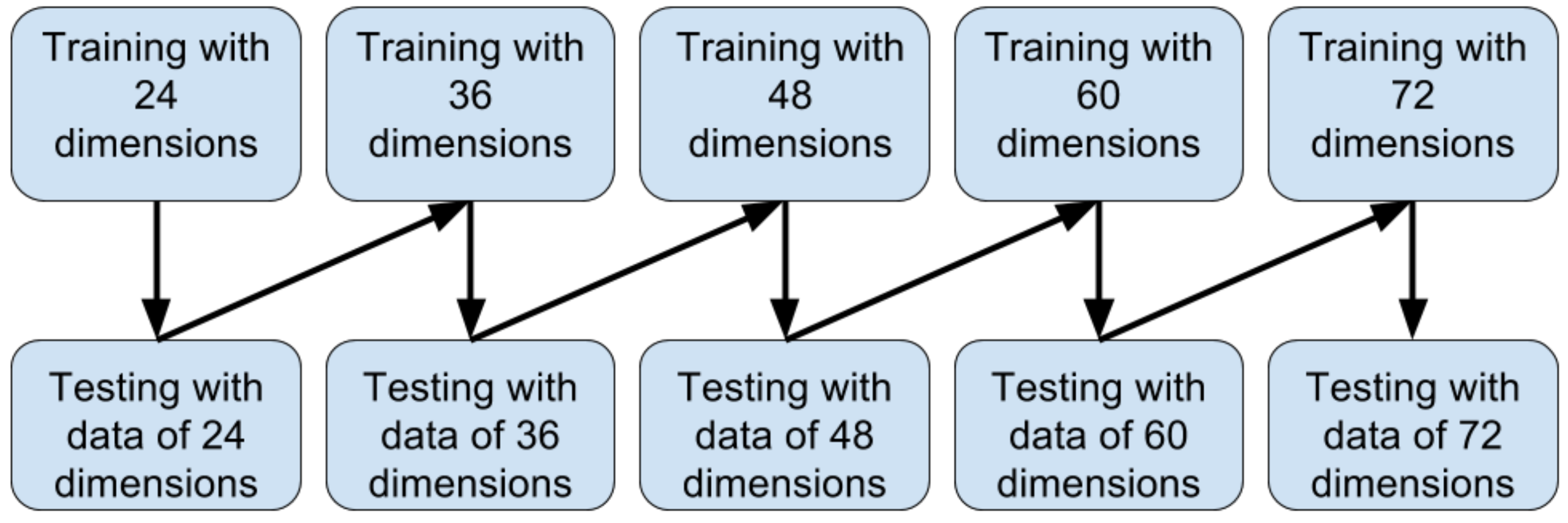}
    \label{fig:modelA}}
    \hfill
    \subfigure[Procedure B]
    {\includegraphics[width=7.9cm]{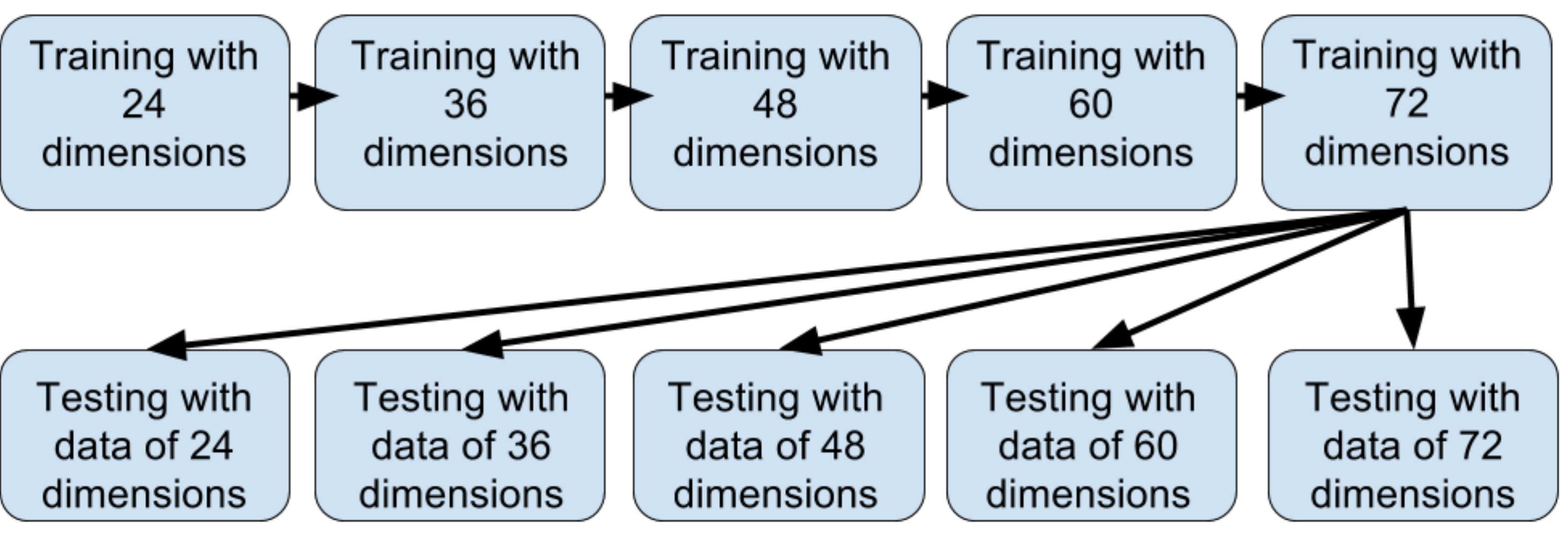}
    \label{fig:modelB}}
    \hfill
    \caption{Training and test procedures designed to verify that VILMAP is able to avoid catastrophic forgetting even after being trained with different dimensions.}
    \label{fig:tests_models}
\end{figure*}

\section{Experiments}\label{sec:experiments}

Three experiments were performed in order to evaluate the proposed approach: the first experiment (\sref{experiment_1}) aims to verify the performance of VILMAP in a standard TSMD problem, the GunPoint dataset \cite{UCRArchive}; the second experiment (\sref{experiment_2}) evaluates the ability of the model in learning sequences of phonemes of increasing size without degrading its performance; Finally, the third experiment (\sref{experiment_3}) compares the performance of the model with the other methods in the literature developed for Word Segmentation.

\subsection{Experiment 1}\label{experiment_1}

GunPoint is a dataset that involves one female actress and one male actor moving they are to form a gun point gesture. Some characteristics of the Gun-Point dataset are: train size = 50; test size = 150; size of each input pattern: 150 points; and number of classes = 2.

The TSMD problem is to identify the two motifs of male and female actors. The main purpose of this experiment is to show that the VILMAP can solve problems of fixed size in the literature, even being created to receive input patterns with different dimensions. In this experiment the parameters were adjusted by trial and error as follows:  ${ a }_{ t } = 0.702,$\quad ${ e }_{ b } = 0.060,$\quad ${ e }_{ n } = 0.247,$\quad $\beta = 0.092,$\quad ${\epsilon}_{ds} = 0.070,$\quad ${ N }_{ max } = 10000 $\quad $and$ \quad $minwd = 0.223$.

\begin{figure*}[!ht]
    \centering
    \hfill
    \subfigure[Mean and std of class 1 (green) with the pattern created (blue)]
    {\includegraphics[width=8.5cm]{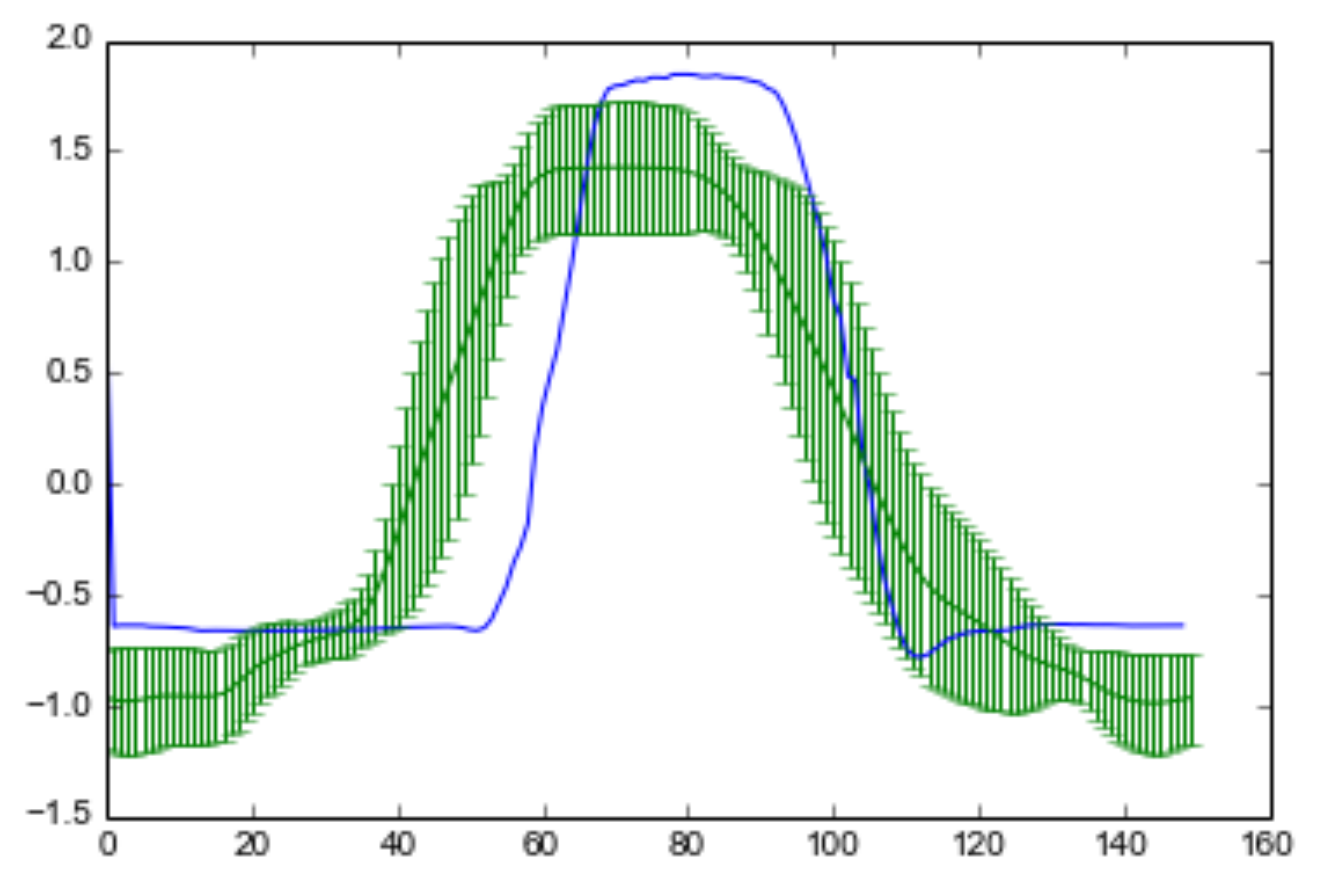}
    \label{fig:tsmd_c1}}
    \hfill
    \subfigure[Mean and std of class 2 (green) with the pattern created (blue)]
    {\includegraphics[width=8.5cm]{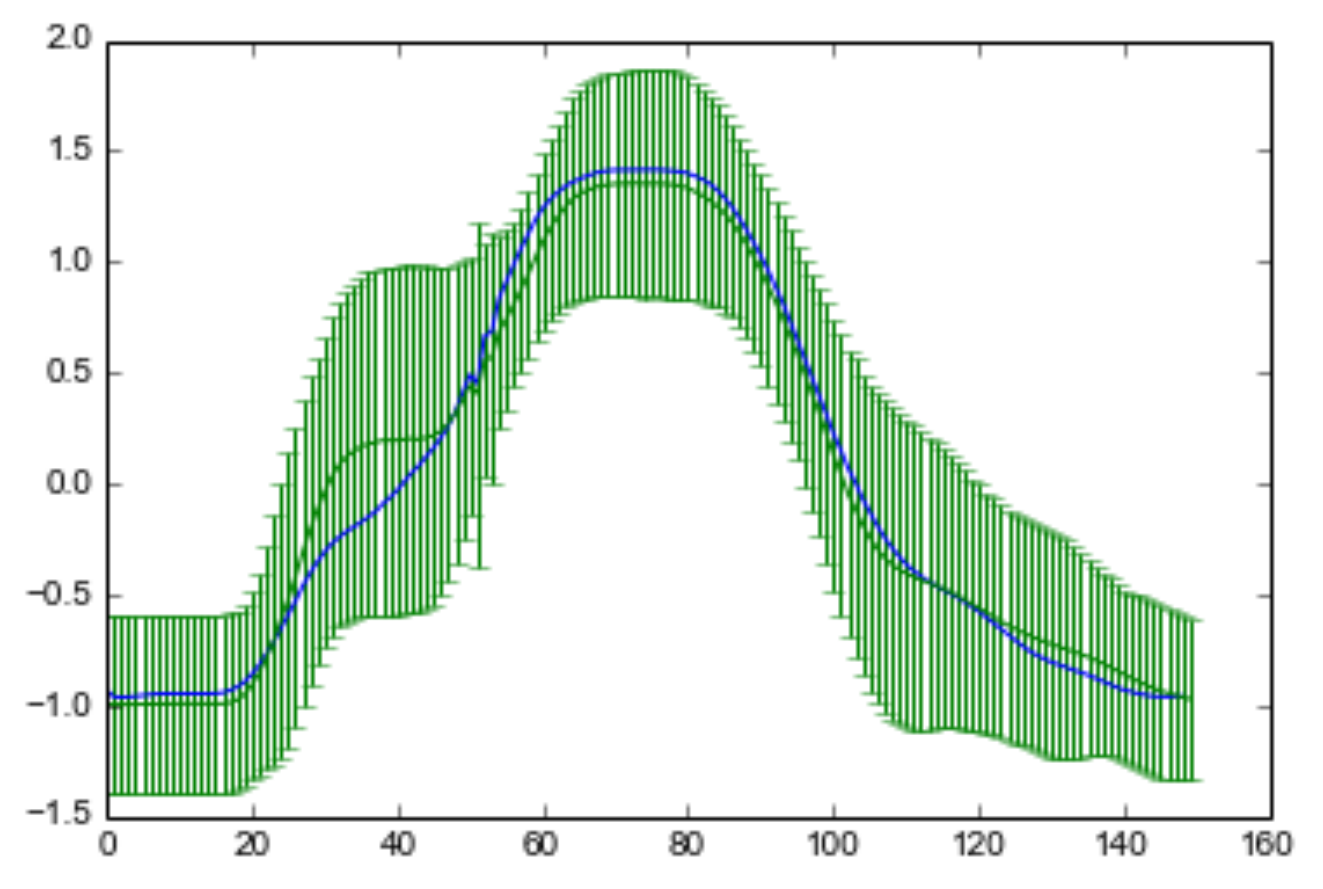}
    \label{fig:tsmd_c2}}
    \hfill
    \caption{Graphical comparison of the Motifs found with the mean and standard deviation of each class of the gun-point time series.}
    \label{fig:tsmd_graphs}
\end{figure*}

With these parameters, we were able to adjust the map to form only two clusters, one for each Motif. \fref{fig:tsmd_graphs} displays a graphical comparison between the Motifs of the Gun-Point dataset found by VILMAP and the Motifs expected in each class with their averages and their standard deviations. In \fref{fig:tsmd_c1} the average of samples from the class is presented with the associated standard deviation (STD), while \fref{fig:tsmd_c2}, the average, and STD for the second class are presented. This result shows that with only one passage through the data, the model was able to find both Motifs correctly, but with a certain displacement in the first one. The second Motif presented a result with the prototype being within the range of standard deviation and very close to average.%With this adjust xxx% of the patters correctly clustered.

\subsection{Experiment 2}\label{experiment_2}
The second dataset is composed of a set of 130 text sentences from the transcripts of TIMIT Acoustic-Phonetic Continuous Speech Corpus \cite{timit} that we extracted from the Natural Language Tool Kit (NLTK) \cite{nltk}.

The input file was translated for a sequence of phonemes using the tool provided by CMU \cite{CMU}, and afterward, each phoneme was translated into a 12-dimensional features vector. The network was trained with the sequence of phonemes features obtained with this procedure (True dataset). To compute the F-Measure, precision and recall, we created a False dataset by consisting phonemes present in the dataset disposed of in random sequences that were not in the True dataset. The obtained False dataset has the same quantity of patterns as the original input file.

In this experiment, two test cycles were generated to answer the following question: Is VILMAP capable of avoiding catastrophic forgetting when trained with input patterns with an increasing number of dimensions? In the procedure illustrated in \fref{fig:modelA}, one training cycle with input patterns of a certain size is followed by a test with the same size, starting with the representation of two phonemes (24 dimensions), increasing by 12 dimensions, up to 72 dimensions (6 phonemes). In \fref{fig:modelB} the test procedure performed after the last training cycle with each of the five input sizes are illustrated.

In order to achieve good results, in this experiment, 100 sets of parameter values were sampled from the ranges in \tref{table:parameters}, according to a Latin Hypercube Sampling (LHS) \cite{helton2005comparison}, and we recorded the best results achieved for each parameter set generated. The LHS is employed to ensure complete coverage of the range of each parameter. The interval of each parameter is divided into 100 intervals of equal probability and a single value is randomly selected from each interval.

\begin{figure*}[!ht]
    \centering
    \hfill
    \subfigure[Procedure A]
    {\includegraphics[width=8.5cm]{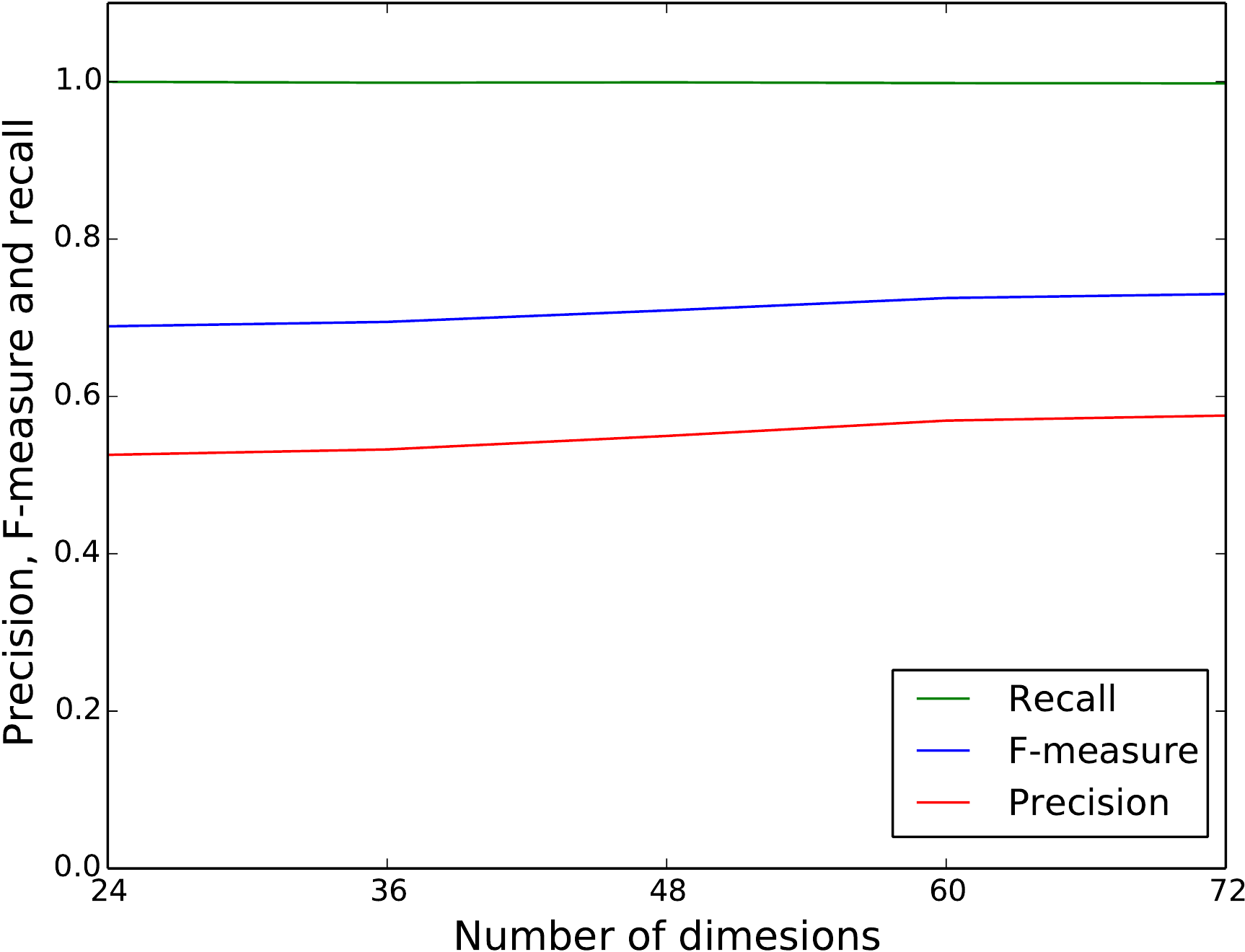}
    \label{fig:modelA_graph}}
    \hfill
    \subfigure[Procedure B]
    {\includegraphics[width=8.5cm]{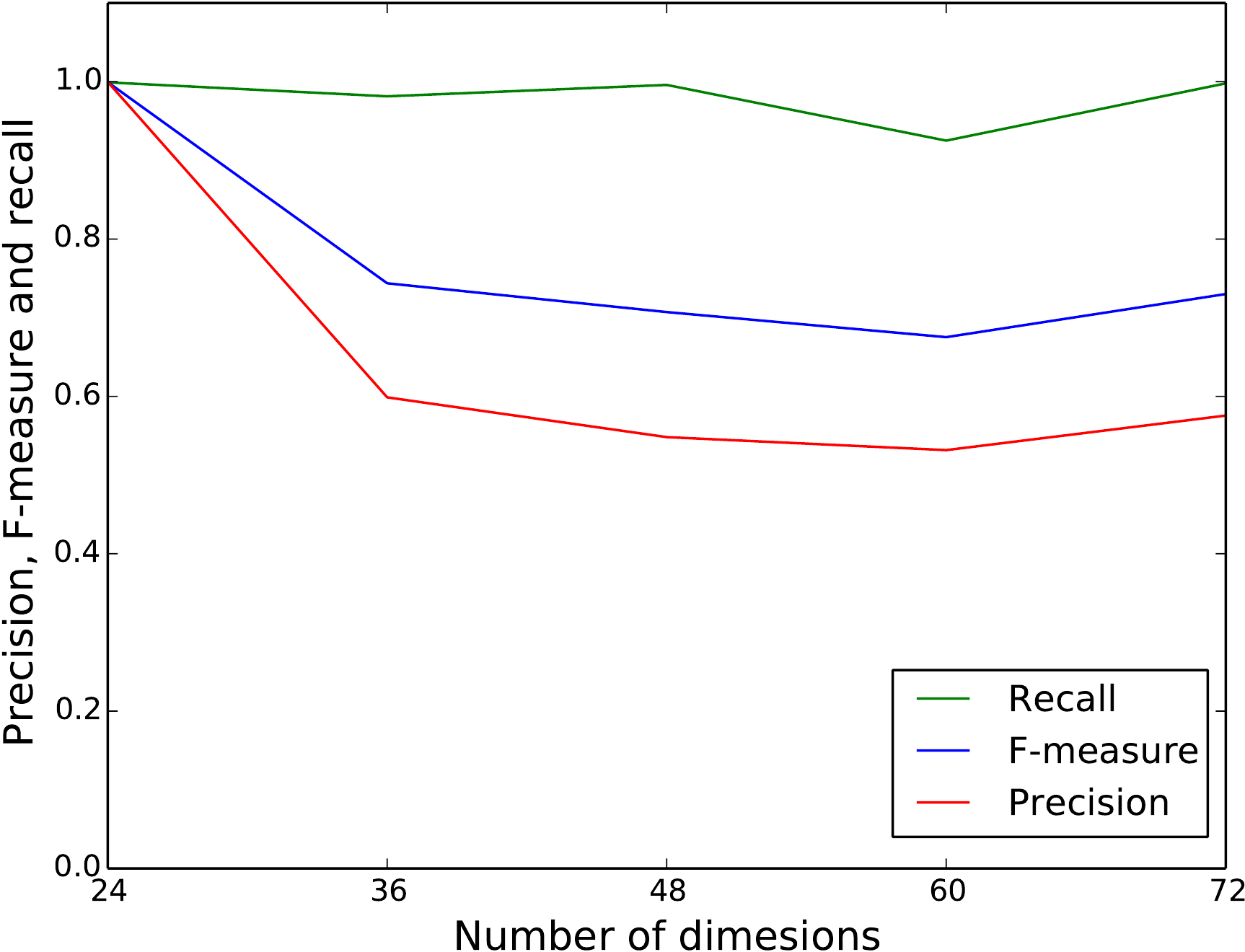}
    \label{fig:modelB_graph}}
    \hfill
    \caption{Graphical Results to compare models A and B.}
    \label{fig:tests_graphs}
\end{figure*}

From the results displayed in \fref{fig:tests_graphs} we can see that in the procedure A, the model achieved an F-Measure of about 70\% a precision of 55\% and a recall of 100\%. By analyzing the \fref{fig:modelB_graph}, we can see that the model did not degrade its performance on lower dimensional datasets after being trained with datasets with a greater number of dimensions. It is worth noting that the performance in the dataset with the lowest dimensions actually increased after training with other datasets. This improvement occurs because nodes of size 24 continue to be updated and recognizing more specific stimuli due to the decrease of its relevances and respective adjustment of the receptive fields.

\subsection{Experiment 3}\label{experiment_3}

The third dataset was the Brent-Siskind corpus \cite{brent_corpus}. This corpus is the largest of the CHIELDS repository \cite{childes} and contains the orthographic transcription of more than 100 hours of recording of 16 English language mothers with children who were between 9 and 15 months old at the time of recording.

In this experiment, we apply VILMAP for segmenting words in transcription of fluent speech. Considering the proposed model does not identify precisely the word boundaries, but is capable of recognizing a word when presented by its inputs, to make it possible to compare the proposed method with the methods presented in \cite{Larsen_Cristia_Dupoux_2017}, we separated non-words from words input patterns. The non-words are caused by the displacements performed on the input data when they are presented to the network since eventually only part of a word is available by the inputs. We present to the network the original words from the database to count the true positives and the false negatives. Finally, as in the previous experiment, we can compute precision, recall and F-measure. The parameter sampling procedure described in the previous experiment was also employed in this experiment.

The results obtained with the proposed method in comparison with the results presented in \cite{Larsen_Cristia_Dupoux_2017} are shown in \tref{table:results}. From this table we can see that VILMAP obtained a relatively good F-measure, only losing for the AGu algorithm; a very good precision, overcoming all other algorithms; and finally, an intermediary recall. We considered this a promising result, since AGu models a learner with infinite memory and a batch process, while our method represents an online learner that passes through the data only once.
\begin{table}[!ht]
\centering
\caption{Results of the word segmentation experiment}
\label{table:results}
\begin{tabular}{l|ccc}
\specialrule{.1em}{.05em}{.05em} 
Algorithm & F-measure & Precision & Recall \\ \hline
VILMAP     & 0.750     & \textbf{0.856}     & 0.667   \\
PUDDLE    & 0.706     & 0.682     & 0.733  \\
DiBS      & 0.236     & 0.234     & 0.240  \\
AGu       & \textbf{0.782}     & 0.787     & \textbf{0.777}  \\
TPs       & 0.468     & 0.432     & 0.512  \\\specialrule{.1em}{.05em}{.05em} 
\end{tabular}
\end{table}

\section{Conclusion and Future Work}\label{sec:conclusion}

This work presented a variable input length self-organizing map, VILMAP, for the problem of motif discovery and word segmentation. The preliminary evaluation presented in Experiment 1, shows that VILMAP can be applied for simple tasks of TSMD. However, a more detailed evaluation of different datasets and conditions is still required.

The results of Experiment 2 showed that the proposed model seems to be able to avoid catastrophic forgetting when the number of dimensions of the input patterns increases with time. This is an interesting and peculiar characteristic of the proposed model not found in other SOM-based models the literature.

Moreover, the results of Experiment 3 show that VILMAP can be a promising candidate for the task of Word Segmentation.

As future work, we intend to explore the capacity of VILMAP of dealing with inputs of variable size to build a recurrent growing self-organizing map, for learning expressions with an increasing level of complexity. In this regard, the fact that VILMAP deals well with such an increasing number of dimensions (up to 72) without degrading its performance on previously learned data is a motivating achievement in this path.

\section*{Acknowledgment}
The authors would like to thank the Brazilian Coordination for the Improvement of Higher Level Personnel (CAPES) for supporting this work.

\bibliographystyle{IEEEtran}
\bibliography{references}

\begin{thebibliography}{10}
\providecommand{\url}[1]{#1}
\csname url@samestyle\endcsname
\providecommand{\newblock}{\relax}
\providecommand{\bibinfo}[2]{#2}
\providecommand{\BIBentrySTDinterwordspacing}{\spaceskip=0pt\relax}
\providecommand{\BIBentryALTinterwordstretchfactor}{4}
\providecommand{\BIBentryALTinterwordspacing}{\spaceskip=\fontdimen2\font plus
\BIBentryALTinterwordstretchfactor\fontdimen3\font minus
  \fontdimen4\font\relax}
\providecommand{\BIBforeignlanguage}[2]{{%
\expandafter\ifx\csname l@#1\endcsname\relax
\typeout{** WARNING: IEEEtran.bst: No hyphenation pattern has been}%
\typeout{** loaded for the language `#1'. Using the pattern for}%
\typeout{** the default language instead.}%
\else
\language=\csname l@#1\endcsname
\fi
#2}}
\providecommand{\BIBdecl}{\relax}
\BIBdecl

\bibitem{torkamani2017survey}
S.~Torkamani and V.~Lohweg, ``Survey on time series motif discovery,''
  \emph{Wiley Interdisciplinary Reviews: Data Mining and Knowledge Discovery},
  vol.~7, no.~2, 2017.

\bibitem{lin2002finding}
J.~Lin, E.~Keogh, S.~Lonardi, and P.~Patel, ``Finding motifs in time series,''
  pp. 53--68, 10 2002.

\bibitem{Kohonen1998}
\BIBentryALTinterwordspacing
T.~Kohonen, ``{The self-organizing map},'' \emph{Neurocomputing}, vol.~21, no.
  1-3, pp. 1--6, nov 1998. [Online]. Available:
  \url{http://linkinghub.elsevier.com/retrieve/pii/S0925231298000307}
\BIBentrySTDinterwordspacing

\bibitem{Rego2007}
R.~L. M.~E. do~Rego, A.~F.~R. Araujo, and F.~B. de~Lima~Neto, ``Growing
  self-organizing maps for surface reconstruction from unstructured point
  clouds,'' in \emph{2007 International Joint Conference on Neural Networks},
  Aug 2007, pp. 1900--1905.

\bibitem{Bassani2015}
\BIBentryALTinterwordspacing
H.~F. Bassani and A.~F.~R. Araujo, ``{Dimension Selective Self-Organizing Maps
  With Time-Varying Structure for Subspace and Projected Clustering},''
  \emph{IEEE Transactions on Neural Networks and Learning Systems}, vol.~26,
  no.~3, pp. 458--471, mar 2015. [Online]. Available:
  \url{http://ieeexplore.ieee.org/document/6803941/}
\BIBentrySTDinterwordspacing

\bibitem{Miikkulainen2005}
R.~Miikkulainen, J.~A. Bednar, Y.~Choe, and J.~Sirosh, \emph{Computational Maps
  in the Visual Cortex}.\hskip 1em plus 0.5em minus 0.4em\relax Springer,
  Janeiro 2005, vol.~1.

\bibitem{Kangas1991}
J.~Kangas, ``Phoneme recognition using time-dependent versions of
  self-organizing maps,'' in \emph{[Proceedings] ICASSP 91: 1991 International
  Conference on Acoustics, Speech, and Signal Processing}, Apr 1991, pp.
  101--104 vol.1.

\bibitem{kohonen1982self}
T.~Kohonen, ``Self-organized formation of topologically correct feature maps,''
  \emph{Biological cybernetics}, vol.~43, no.~1, pp. 59--69, 1982.

\bibitem{keogh2011curse}
E.~Keogh and A.~Mueen, ``Curse of dimensionality,'' in \emph{Encyclopedia of
  machine learning}.\hskip 1em plus 0.5em minus 0.4em\relax Springer, 2011, pp.
  257--258.

\bibitem{reviewClustering2004}
\BIBentryALTinterwordspacing
L.~Parsons, E.~Haque, and H.~Liu, ``Subspace clustering for high dimensional
  data: A review,'' \emph{SIGKDD Explor. Newsl.}, vol.~6, no.~1, pp. 90--105,
  Jun. 2004. [Online]. Available:
  \url{http://doi.acm.org/10.1145/1007730.1007731}
\BIBentrySTDinterwordspacing

\bibitem{clusteringSurvey2009}
\BIBentryALTinterwordspacing
H.-P. Kriegel, P.~Kr\"{o}ger, and A.~Zimek, ``Clustering high-dimensional data:
  A survey on subspace clustering, pattern-based clustering, and correlation
  clustering,'' \emph{ACM Trans. Knowl. Discov. Data}, vol.~3, no.~1, pp.
  1:1--1:58, Mar. 2009. [Online]. Available:
  \url{http://doi.acm.org/10.1145/1497577.1497578}
\BIBentrySTDinterwordspacing

\bibitem{DSSOM}
H.~F. Bassani and A.~F. Ara{\'u}jo, ``Dimension selective self-organizing maps
  for clustering high dimensional data,'' in \emph{Neural Networks (IJCNN), The
  2012 International Joint Conference on}.\hskip 1em plus 0.5em minus
  0.4em\relax IEEE, 2012, pp. 1--8.

\bibitem{nunthanid2011discovery}
P.~Nunthanid, V.~Niennattrakul, and C.~A. Ratanamahatana, ``Discovery of
  variable length time series motif,'' in \emph{Electrical
  Engineering/Electronics, Computer, Telecommunications and Information
  Technology (ECTI-CON), 2011 8th International Conference on}.\hskip 1em plus
  0.5em minus 0.4em\relax IEEE, 2011, pp. 472--475.

\bibitem{UCRArchive}
Y.~Chen, E.~Keogh, B.~Hu, N.~Begum, A.~Bagnall, A.~Mueen, and G.~Batista, ``The
  ucr time series classification archive,'' July 2015.

\bibitem{mueen2014}
\BIBentryALTinterwordspacing
A.~Mueen, ``Time series motif discovery: dimensions and applications,''
  \emph{Wiley Interdisciplinary Reviews: Data Mining and Knowledge Discovery},
  vol.~4, no.~2, pp. 152--159, 2014. [Online]. Available:
  \url{http://dx.doi.org/10.1002/widm.1119}
\BIBentrySTDinterwordspacing

\bibitem{kohonen1990self}
T.~Kohonen, ``The self-organizing map,'' \emph{Proceedings of the IEEE},
  vol.~78, no.~9, pp. 1464--1480, 1990.

\bibitem{saffran1996word}
J.~R. Saffran, E.~L. Newport, and R.~N. Aslin, ``Word segmentation: The role of
  distributional cues,'' \emph{Journal of memory and language}, vol.~35, no.~4,
  pp. 606--621, 1996.

\bibitem{Correa2007}
J.~Correa and J.~E. Dockrell, ``Unconventional word segmentation in brazilian
  children's early text production,'' \emph{Reading and Writing}, vol.~20,
  no.~8, pp. 815--831, Nov 2007.

\bibitem{Larsen_Cristia_Dupoux_2017}
\BIBentryALTinterwordspacing
E.~Larsen, A.~Cristia, and E.~Dupoux, ``Relating unsupervised word segmentation
  to reported vocabulary acquisition,'' Jun 2017. [Online]. Available:
  \url{osf.io/wa6tq}
\BIBentrySTDinterwordspacing

\bibitem{daland2011learning}
R.~Daland and J.~B. Pierrehumbert, ``Learning diphone-based segmentation,''
  \emph{Cognitive science}, vol.~35, no.~1, pp. 119--155, 2011.

\bibitem{saffran1996statistical}
J.~R. Saffran, R.~N. Aslin, and E.~L. Newport, ``Statistical learning by
  8-month-old infants,'' \emph{Science}, vol. 274, no. 5294, pp. 1926--1928,
  1996.

\bibitem{monaghan2010words}
P.~Monaghan and M.~H. Christiansen, ``Words in puddles of sound: Modelling
  psycholinguistic effects in speech segmentation,'' \emph{Journal of child
  language}, vol.~37, no.~3, pp. 545--564, 2010.

\bibitem{johnson2007adaptor}
M.~Johnson, T.~L. Griffiths, and S.~Goldwater, ``Adaptor grammars: A framework
  for specifying compositional nonparametric bayesian models,'' in
  \emph{Advances in neural information processing systems}, 2007, pp. 641--648.

\bibitem{timit}
J.~S. Garofolo, L.~F. Lamel, W.~M. Fisher, J.~G. Fiscus, D.~S. Pallett, N.~L.
  Dahlgren, and V.~Zue, ``Timit acoustic-phonetic continuous speech corpus,
  1993,'' \emph{Linguistic Data Consortium, Philadelphia}.

\bibitem{nltk}
\BIBentryALTinterwordspacing
S.~Bird and E.~Loper, ``Nltk: the natural language toolkit,'' in
  \emph{Proceedings of the ACL 2004 on Interactive poster and demonstration
  sessions}.\hskip 1em plus 0.5em minus 0.4em\relax Association for
  Computational Linguistics, 2004, p.~31. [Online]. Available:
  \url{http://www.nltk.org}
\BIBentrySTDinterwordspacing

\bibitem{CMU}
CMU, ``The carnegie mellon university pronouncing dictionary - a
  machine-readable pronunciation dictionary for north american english.
  on-line.'' 2011.

\bibitem{helton2005comparison}
J.~C. Helton, F.~Davis, and J.~D. Johnson, ``A comparison of uncertainty and
  sensitivity analysis results obtained with random and latin hypercube
  sampling,'' \emph{Reliability Engineering \& System Safety}, vol.~89, no.~3,
  pp. 305--330, 2005.

\bibitem{brent_corpus}
\BIBentryALTinterwordspacing
M.~R. Brent and J.~M. Siskind, ``The role of exposure to isolated words in
  early vocabulary development,'' in \emph{Cognition}, vol.~81, 2001, pp.
  31--44. [Online]. Available:
  \url{https://childes.talkbank.org/access/Eng-NA/Brent.html}
\BIBentrySTDinterwordspacing

\bibitem{childes}
B.~MacWhinney, \emph{The childes project: Tools for analyzing talk.}\hskip 1em
  plus 0.5em minus 0.4em\relax Psychology Press, 2000, vol.~2.

\end{thebibliography}

\end{document}